\documentclass[runningheads]{llncs}

\usepackage{eccv}

\usepackage{eccvabbrv}

\usepackage{graphicx}
\usepackage{booktabs}
\usepackage[rightcaption]{sidecap}
\usepackage{siunitx}

\usepackage[accsupp]{axessibility}  
\usepackage{tikz}
\usepackage{multirow, multicol}
\usepackage{comment}
\usepackage{amsmath,amssymb} 
\usepackage{color}
\usepackage{pifont}

\usepackage{hyperref}

\usepackage{orcidlink}

\begin{document}

\title{MERLiN: Single-Shot Material Estimation and Relighting for Photometric Stereo} 

\titlerunning{MERLiN}

\author{Ashish Tiwari\inst{1}\orcidlink{0000-0002-4462-6086} \and
Satoshi Ikehata\inst{2}\orcidlink{0000-0002-6061-7956} \and
Shanmuganathan Raman\inst{1}\orcidlink{0000-0003-2718-7891}}

\authorrunning{A.~Tiwari et al.}

\institute{Indian Institute of Technology Gandhinagar, Gujarat, India \\ 
\email{\{ashish.tiwari, shanmuga\}@iitgn.ac.in}\\ \and
National Institute of Informatics, Tokyo, Japan\\
\email{sikehata@nii.ac.jp}}

\maketitle

\begin{abstract}
  Photometric stereo typically demands intricate data acquisition setups involving multiple light sources to recover surface normals accurately. In this paper, we propose MERLiN, an attention-based hourglass network that integrates single image-based inverse rendering and relighting within a single unified framework. We evaluate the performance of photometric stereo methods using these relit images and demonstrate how they can circumvent the underlying challenge of complex data acquisition. Our physically-based model is trained on a large synthetic dataset containing complex shapes with spatially varying BRDF and is designed to handle indirect illumination effects to improve material reconstruction and relighting. Through extensive qualitative and quantitative evaluation, we demonstrate that the proposed framework generalizes well to real-world images, achieving high-quality shape, material estimation, and relighting. We assess these synthetically relit images over photometric stereo benchmark methods for their physical correctness and resulting normal estimation accuracy, paving the way towards single-shot photometric stereo through physically-based relighting. This work allows us to address the single image-based inverse rendering problem holistically, applying well to both synthetic and real data and taking a step towards mitigating the challenge of data acquisition in photometric stereo.
\end{abstract}

\section{Introduction}
\label{sec:intro}

Photometric stereo \cite{woodham1980photometric} plays a pivotal role in 3D reconstruction, surface analysis, and material recovery. By analyzing an object's appearance under multiple illumination conditions, it infers per-pixel surface normals. It directly extends to applications such as quality control, industrial inspection, medical imaging, cultural heritage preservation, and robotics, to name a few. However, despite its utility, photometric stereo encounters several challenges that constrain its applicability and accuracy in real-world scenarios. 

One significant challenge lies in the complexity of data acquisition, which often demands carefully orchestrated setups involving controlled lighting environments and precise calibration procedures. Due to practical constraints such as time, cost, and equipment limitations, it is often infeasible to exhaustively sample the entire space of possible lighting configurations. As a result, the acquired dataset may not sufficiently cover all relevant lighting variations, leading to incomplete or inaccurate surface reconstructions. 

\textbf{Key Questions.} (a) \textit{Can we leverage the advancements in deep learning research to generate differently illuminated images?} Image relighting has been addressed from various perspectives using deep learning. One stream of works  \cite{isola2017image, zhu2017unpaired, xu2018deep, li2018cgintrinsics, li2018learning, sang2020single} includes the use of convolutional neural networks (CNN), while the other stream of works \cite{wu2023nefii, srinivasan2021nerv, xu2023renerf, li2023relit, zhang2021neural} is based on neural radiance fields (NeRF) \cite{mildenhall2021nerf} for relighting and material estimation. The NeRF-based methods have extensively improved the relighting results. However, they rely on multiple calibrated images and lengthy per-scene optimization. Interestingly, CNN-based approaches have achieved relighting in a feed-forward manner from a sparse set of views (as few as one). Initially, works like \cite{isola2017image, zhu2017unpaired, xu2018deep} modeled relighting as an image-to-image translation task where a CNN can be trained with one or more images and novel lighting as input to generate the relit targets. 

(b) \textit{Do these synthesized images always guarantee the physical correctness of the relit images?} The image-based relighting methods often produce images that are not physically meaningful because the images may ``appear'' perceptually realistic even if the underlying shape and material parameters deviate from being physically correct. Physically correct image relighting fundamentally demands an in-depth understanding of geometry, material properties, and illumination. The challenge is more compounded when addressing objects of diverse textures and reflectance properties since these elements interact in complex ways. A stream of works \cite{li2018cgintrinsics, li2018learning, sang2020single} perform relighting through intrinsic parameter estimation. The relighting is performed either via a neural network \cite{sang2020single} or through an in-network rendering layer simulating a specific BRDF model \cite{li2018learning}. Such an approach offers better controllability and editability of scene parameters. Furthermore, global illumination plays a vital role in physical plausibility. While works like \cite{li2018learning} consider global illumination effects due to indirect light bounces, most other CNN-based methods have resorted to direct illumination.

\begin{figure}[t]
	\centering    
	\includegraphics[width=\linewidth]{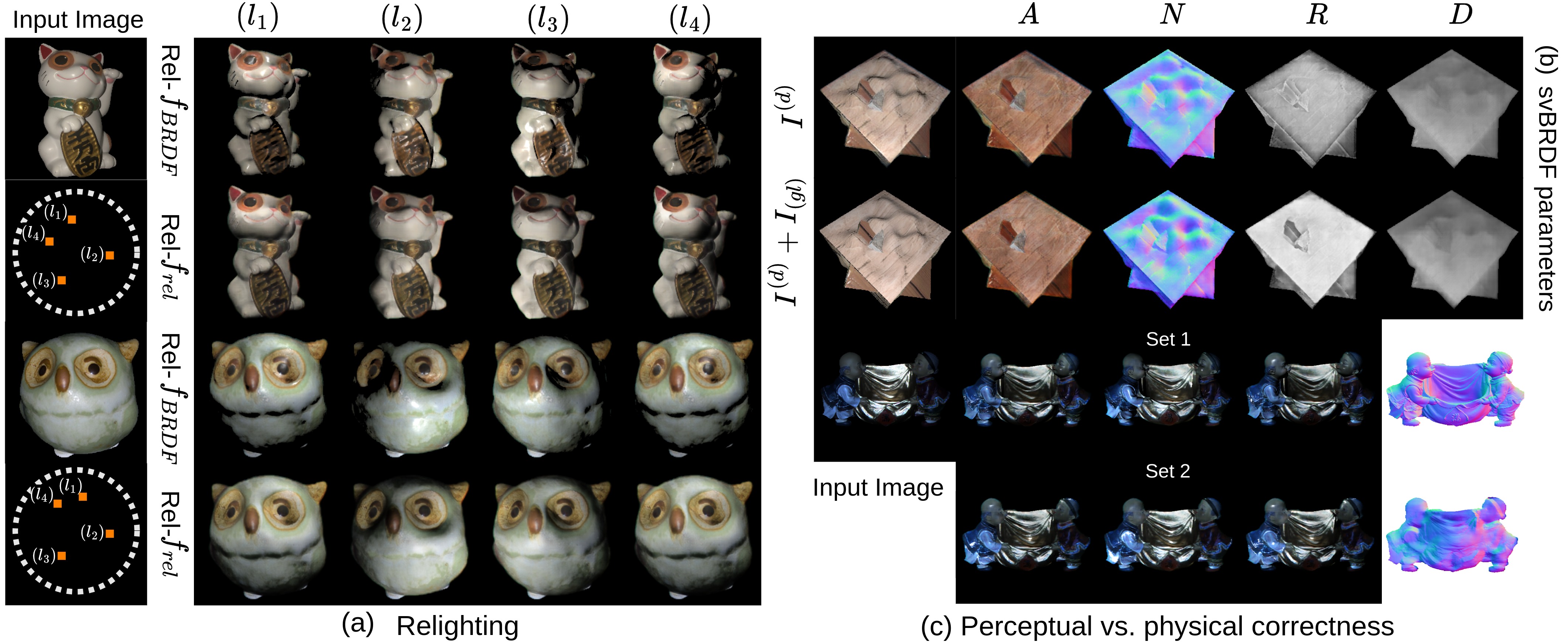}
	\caption{ (a) Effect of relighting (under four light positions $l_{1}, l_{2}, l_{3}, l_{4}$) through BRDF rendering layer $f_{BRDF}$ and neural network $(f_{rel})$, (b) Effect of training with direct (top) vs global illumination (bottom) images. The estimated normals without global illumination are flattened and produce brighter albedo (top). (c) Two different sets of perceptually similar images with different underlying normals maps.}
	\label{fig:abl}

\end{figure}

(c) \textit{How can we validate the physical correctness of these relit images?} Interestingly, photometric stereo itself offers a solution. As shown in Figure \ref{fig:abl} (c), two sets 
of images with similar perceptual fidelity can result in widely different normals. Such \textit{``perceptually-correct physically-incorrect''} images fail to generate correct normal estimates through the photometric stereo. Therefore, one could also evaluate the relit images by measuring the performance on the photometric stereo.

\textbf{Key Ideas and Contributions} The following are the key contributions to address the aforementioned observations.

(i) We propose a physically-based global illumination-aware deep network, called MERLiN - \underline{M}aterial \underline{E}stimation and \underline{R}e\underline{Li}ghting \underline{N}etwork, to estimate spatially varying bidirectional reflectance distribution function (svBRDF) parameters such as diffuse albedo, normal, depth, and specular roughness) and jointly perform relighting through a single image. We perform relighting through estimated image intrinsics and learn the complex relationship between appearance and lighting. The joint learning allows the network to simulate a physically-based rendering process \cite{sang2020single} and ensures that the relit images are close to their real counterparts.

(ii) We validate the physical correctness of the relit images through existing photometric stereo benchmarks and compare the accuracy of normal estimation using the relit images and their real counterparts. 
This way, we take a step towards addressing photometric stereo from a single image via image-based relighting. 



\section{Related Work}
\label{sec:related_work}

\textbf{Shape and Material Estimation.} Several deep learning frameworks have been designed for the inverse problem over indoor \cite{sengupta2019neural} and outdoor \cite{yu2021outdoor} scenes for material recognition \cite{bell2015material} and estimation \cite{meka2018lime}, reflectance maps extraction \cite{rematas2016deep}, surface appearance recovery \cite{li2017modeling}, normal and depth estimation \cite{eigen2015predicting}. Others assume a specific class of objects, such as faces \cite{shu2017neural,zhou2019deep,sun2019single} or near planar surfaces \cite{aittala2016reflectance, li2017modeling, li2018materials} for shape and reflectance recovery. Further, some methods apply to images captured through smartphones \cite{li2018learning, sang2020single} and a few on in-the-wild images \cite{wimbauer2022rendering}. The ill-posed nature of the problem, especially using a single image, demands more labeled training data for ground supervision. While Li \emph{et al.} \cite{li2017modeling} leverage the appearance information embedded in unlabeled images of spatially varying materials to self-augment the training process, primarily to reduce the amount of required labeled training data, The authors in \cite{deschaintre2018single,li2018learning, li2018materials} train CNNs to regress svBRDF and surface normal using in-network rendering to provide additional supervision during training. However, Sang \emph{et al.} \cite{sang2020single} use CNNs to jointly estimate svBRDF parameters and perform relighting with a single image. While Yi \emph{et al.} \cite{yi2023weakly} use differently lit images during training, they perform single image-based inference during test time. However, they use an off-the-shelf network to remove specularities before performing intrinsic decomposition. In another approach,  Wimbauer \emph{et al.} \cite{wimbauer2022rendering} use the priors learned by other networks to aid in the shape and material reflectance. Interestingly, different works have addressed inverse rendering in different flavors, such as intrinsic image decomposition \cite{li2018cgintrinsics,liu2020unsupervised} or specularity removal \cite{yamamoto2019general,shi2017learning} or surface normal estimation \cite{li2018learning, sang2020single} or even through photometric stereo \cite{li2022neural, li2022self, tiwari2022deepps2}.\\ 
\textbf{Image Relighting.} Image-based relighting has been approached through an image translation perspective \cite{isola2017image, zhu2017unpaired}. Several other methods \cite{xu2018deep, yi2023weakly} have used a sparse set of multiple images for relighting through CNNs. While single image-based relighting is highly ill-posed, methods like \cite{zhou2019deep} have performed relighting over facial images. Others have performed relighting either using an in-network rendering layer \cite{li2018learning} or training a relighting network jointly with intrinsic parameter estimation \cite{sang2020single} from a single image. In this work, we follow the paradigm of \cite{li2018learning, sang2020single} with two important differences - (i) a single-stage network (in contrast to their cascaded network design) with (ii) an in-network global illumination handling for joint material estimation and relighting from a single image allowing us to better model the shape, illumination, and appearance dependencies. While \cite{li2018learning} considers global illumination effects through a cascaded CNN, their training is not end-to-end with svBRDF estimation. Specifically, they train the global illumination network separately to estimate second and third light bounce images, given the first bounce image. On the contrary, we use a single-stage global illumination network to perform end-to-end training with material estimation and relighting.\\
\textbf{Photometric stereo.} Earlier to photometric stereo, Shape from Shading  (SfS) \cite{horn1970shape, horn1989shape} methods were proposed to reconstruct shape from single images captured under calibrated illumination, though they usually assume Lambertian reflectance \cite{johnson2011shape}. Later, they were extended to arbitrary shapes and reflectance under known natural illumination \cite{oxholm2015shape}. However, due to the severely ill-posed nature of the problem, researchers incorporated multiple images under different lightings for shape estimation and addressed Photometric Stereo. Several methods \cite{woodham1980photometric, goldman2009shape, li2022neural, li2022self, tiwari2022deepps2, ikehata2023scalable} with the help of meaningfully curated deep learning-based architectures have been used to recover shape, BRDF material, and lighting by generally solving an optimization problem using multiple images of a scene captured under different lighting conditions and/or from multiple viewpoints.  However, acquiring these multiple images with controlled lighting for either training or inference is tedious and challenging to apply to objects under arbitrary illumination. Some methods like \cite{tiwari2022lerps} and \cite{tiwari2022deepps2} have performed photometric stereo through relighting in supervised and self-supervised manner, respectively, using one and two images as input during inference, but multiple images (one or two at a time) for training. Others have performed inverse rendering for photometric stereo in a self-supervised manner \cite{kaya2021uncalibrated, li2022neural, li2022self}. Interestingly, none of the existing works have demonstrated the use of relit images to solve photometric stereo. We take the first step towards photometric stereo through single image-based material estimation and relighting. Our attempt can also be viewed as a bridge between shape from shading \cite{horn1970shape, horn1989shape} and photometric stereo \cite{woodham1980photometric}.

\begin{figure}[t]
	\centering
	\includegraphics[width=\textwidth]{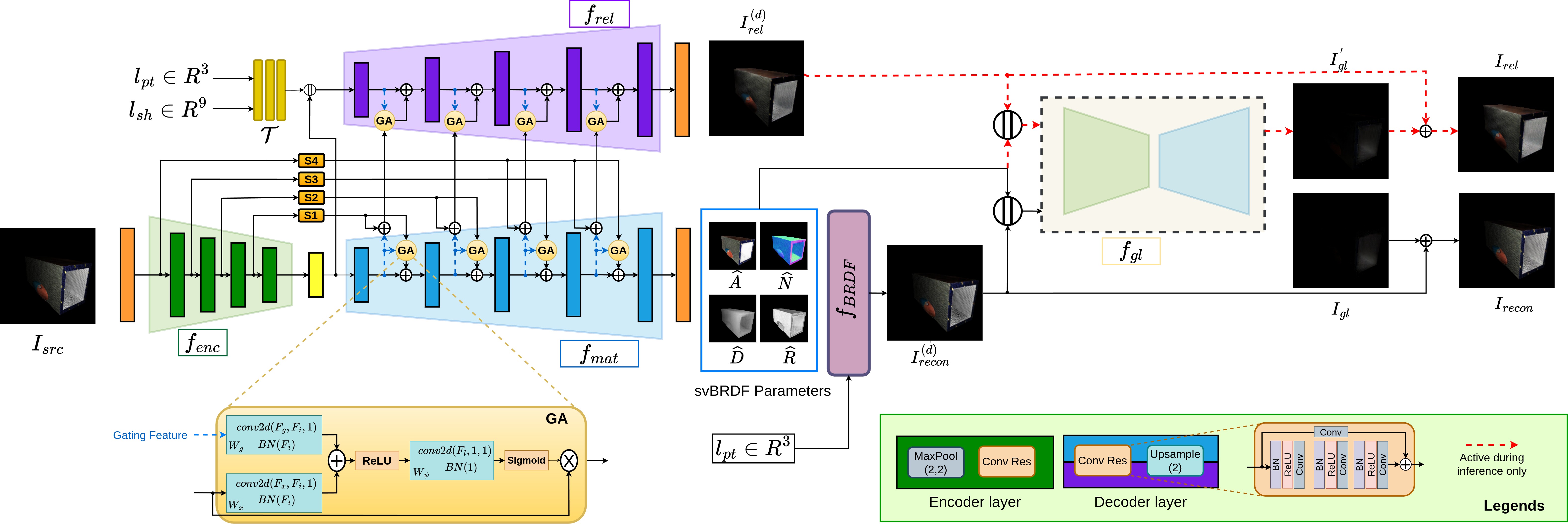}
	\caption{The proposed framework for single image-based $(I_{src})$ svBRDF estimation $(\widehat{A}, \widehat{N}, \widehat{D}, \widehat{R})$ and relighting $(I_{ref})$. The design of the encoder and decoder of the \textit{global illumination network} ($f_{gl}$) is the same as $f_{enc}$ and $f_{inv\_dec}$, respectively. The superscript $(d)$ and subscript $(gl)$ represent the direct and indirect illumination, respectively. S1-S4 are residual skip connections.}
	\label{fig:bd}

\end{figure}

\section{Method}
\label{sec:method}
\textbf{Objective:} Given a single image of an object under point and/or uncontrolled environment illumination and target lighting, we would like to first estimate four svBRDF parameters - diffuse albedo $(A)$, normal $(N)$, depth $(D)$, and roughness $(R)$ and relight the image under target lighting. Further, we want to use multiple different relit images as input to the two photometric stereo methods \cite{lichy2022fast, ikehata2023scalable} to evaluate the surface normal estimates and validate the physical correctness of the relit images. 

\subsection{\underline{M}aterial \underline{E}stimation and \underline{R}e\underline{Li}ghting \underline{N}etwork (MERLiN)}
We propose a single-stage attention-guided convolutional neural network, called MERLiN, for joint material estimation and relighting from a single image. The network consists of one shared encoder and two decoders, one each for svBRDF estimation and relighting. The skip connections among the encoder and decoders are used for feature sharing across different layers. The design of MERLiN is inspired by the hourglass networks \cite{newell2016stacked, yang2017stacked} that are well known for hierarchical feature learning and multi-scale processing over different tasks. They have also been applied to inverse rendering \cite{yu2019inverserendernet} and relighting \cite{zhou2019deep} tasks.

As per Figure \ref{fig:bd}, let us consider an image $I_{src}$ (multiplied with the binary mask) as the input to the encoder $f_{enc}$. 

\textbf{svBRDF Estimation.} Consider a set of features extracted by the encoder as $Z_{enc}$ \ie, $Z_{enc} = f_{enc}(I_{src}, M)$. These features are passed to the \textit{material decoder} $f_{mat}$ to obtain diffuse albedo $(\widehat{A})$, roughness $(\widehat{R})$, surface normal $(\widehat{N})$, and depth $(\widehat{D})$ jointly rather than independently, such that the following holds.
\begin{equation}
    \centering
    \widehat{A}, \widehat{R}, \widehat{N}, \widehat{D} = f_{mat}(Z_{enc})
    \label{eq:1}
\end{equation}

The associated skip connections are combined at the respective layer in the decoder. We use only a single decoder to model correlations between the object's shape and material. Such a design has fewer parameters and offers faster runtime speed than the existing cascaded designs \cite{li2018learning, sang2020single}.

\textbf{Feature fusion through attention gating.} N\"aively combining the features from the skip connections with the decoder features at respective scales leads to poor results (see Table \ref{tab:1}-IDs $1$ and $2$), primarily due to the underlying redundancies and noise in the skip connections. Therefore, we adopt the attention-gating mechanism of \cite{oktay2018attention} for feature fusion. The information extracted from the coarse scale in the decoder is used as a gating signal to disambiguate irrelevant and noisy responses in skip connections. Such an approach also captures local and global effects such as surface roughness, textures, light intensity fall-off, and specular regions to better model the light interaction with surfaces for joint material estimation and relighting.

\textbf{Image Reconstruction.} To validate the correctness of the estimated intrinsic parameters, we reconstruct the input image through the rendering layer following the microfacet BRDF model \cite{karis2013real, hill2020physically}. Given the diffuse albedo (A), specular roughness (R), normals (N), and depth (D) along with $l$, $v$, and $h$ being the light direction, viewing angle, and the half-angle between them, one can easily render the image $I^{(d)}$ as per Equation \ref{eq:2}.
\begin{equation}
    \centering
    I^{(d)} = \frac{1}{D^{2}} \cdot f_{BRDF}(A,R,N,D,l,v)
    \label{eq:2}
\end{equation}
Here, superscript $d$ indicates the image generated under direct illumination, and $1/D^{2}$ accounts for the light fall-off. The BRDF $(f_{BRDF})$ can be characterized as per Equation \ref{eq:3}.
\begin{equation}
    \centering
    f_{BRDF}(A,N,R,D,l,v) = \frac{A}{\pi} +  \frac{\widetilde{M}_{f}(h, R)\widetilde{F}(v,h)\widetilde{G}(l,v,h,R)}{4(N\cdot l)(N \cdot v)}
    \label{eq:3}
\end{equation}
Here, $\widetilde{M}_{f}(h,R)$, $\widetilde{F}(v,h)$, and $\widetilde{G}(l,v,h,R)$ are the microfacet distribution, Fresnel, and geometry term. A detailed description of the BRDF model is provided in the supplementary material.

\textbf{Global Illumination.} Incorporating global illumination is crucial for relighting and rendering, especially over intricate shapes with complex material and reflectance such as high specularities or glossiness. Several existing works on material capture \cite{li2018materials, sang2020single, yu2019inverserendernet} and photometric stereo \cite{li2022self, lichy2021shape, tiwari2022lerps} do not explicitly model global illumination effects such as indirect illumination and inter-reflections. While a prior work\cite{li2018learning} has considered global illumination, it trained a two-stage cascaded network separately from the BRDF estimation network. However, we train a single-stage global illumination network along with the BRDF estimation network. The end-to-end learning allows for better co-guidance among global illumination network and material decoder to produce better results. We predict the combined indirect illumination across multiple light bounces instead of modeling individual light bounces, as in \cite{li2019learning}. The global illumination network $(f_{gl})$ design is similar to the encoder-decoder framework described earlier. 
The output of the rendering layer $I^{(d)}$ (direct illumination) along with estimated intrinsic parameters $(\widehat{A}, \widehat{R}, \widehat{N}, \widehat{D})$ is fed to the network and produces the residual image $I_{gl}$ which when combined with the $I^{(d)}$ yields the final image $I$ with global illumination effects. The residual image is expected to capture the energy contained in higher-order light bounces. Equation \ref{eq:5} describes the image formation with global illumination.
\begin{equation}
    \centering    
    \begin{split}
        I = I^{(d)} + I_{gl} \text{   s.t.  } I_{gl} = f_{gl}(I^{(d)}, \widehat{A}, \widehat{R}, \widehat{N}, \widehat{D})
    \end{split}        
    \label{eq:5}
\end{equation}
Interestingly, as shown in Figure \ref{fig:abl} (b), the network trained with direct lighting only predicts brighter diffuse albedo and flattened normals when evaluated on images with indirect lighting, which aligns with observation in \cite{chandraker2005reflections, li2018learning}. Therefore, we train our network over images with global illumination.

\textbf{Relighting.} We explore two ways of relighting - one through a physically-based rendering based on predicted BRDF (Rel-$f_{BRDF})$ and the other through a CNN-based decoder (Rel-$f_{rel}$).

\textbf{(a)} Rel-$f_{BRDF}$. One way to relight is to use the estimated BRDF parameters and directly render the image under arbitrary target lighting $l_{tar}$ through a BRDF model, as described in Equation \ref{eq:2}. Such an approach ensures that relighting explicitly considers the physical plausibility of the intrinsic parameters to better model the global effects, such as specularities and light fall-off. While this approach would generate images under direct illumination, we use the global illumination network (trained on direct illumination images) to obtain the indirect illumination effects and incorporate them for obtaining physically plausible and visually realistic relit images. 

\textbf{(b)} Rel-$f_{rel}$. Another approach is to train a relighting network with material estimation jointly. The features from the encoder $Z_{enc}$ along with the target lighting information $l_{tar}$ is passed through the \textit{relighting decoder} $f_{rel}$ to generate the image $I_{rel}$ relit under target lighting. The skip connections from the encoder and \textit{material decoder} are combined with the \textit{relighting decoder} at the respective scale through the gated-attention mechanism, such that the Equation \ref{eq:4} holds.
\begin{equation}
    \centering
    I_{rel}^{(d)} = f_{rel}(Z_{enc}, Z_{mat}, \mathcal{T}(l_{tar}))
    \label{eq:4}
\end{equation}
Here, $\mathcal{T}$ is a lighting encoder comprising three MLPs that takes the lighting vector as input. The lighting vector could be a $3 \times 1$ vector for a point or directional light and a $9 \times 1$ vector representing the spherical harmonic coefficients for the environmental illumination. While considering both point and environment lighting, we concatenate both the lighting vectors before feeding to the lighting encoder $\mathcal{T}$. Interestingly, the bidirectional connections across the two decoders allow for better learning of correlated information across two different but related tasks of inverse rendering and relighting. Additionally, it provides additional supervision for inverse parameter estimation guided by expected consistency in the relit images. The relighting decoder generates images under direct illumination during training. However, during inference, the output of the relighting decoder $I_{rel}^{(d)}$ is passed through the trained global illumination network to infer the higher-order light bounce image $I_{gl}$ and finally combined to obtain the final image $I_{rel} = I_{rel}^{(d)} + I_{gl}$ with global illumination effect.

\subsection{Training Details}
We train MERLiN over a large synthetic dataset proposed by \cite{li2018learning} that contains BRDF parameters and images under point and environment lighting. It contains images under camera-co-located near-field point lighting, \ie, $l_{pt}= [0,0,0]$ and object placed at $[0,0,-1]$) \footnote{The dataset uses a camera-centric coordinate system with the camera at the origin and $x,y,z$ directions correspond to $u,v,d$.} and environment lighting ($l_{sh}$) represented by $9$ SH coefficients per color channel. There are three separate point light images, one for the direct component $(I_{pt}^{(d)})$ and two for subsequent light bounces each (that we combine together to obtain a single image $I_{gl}$), and one image under environment lighting $I_{env}$ per object. We train MERLiN under two settings: (A) point lighting and (B) point + environment lighting. For setting (A), we consider one point light image such that $I_{src} = I_{pt}^{(d)}+I_{gl}$ with the direct component $I^{(d)} = I_{pt}^{(d)}$, global illumination effects $(I_{gl})$, and target lighting $l_{tar} = l_{pt}$. For setting (B), we consider $I_{src} = (I_{pt}^{d} + I_{env}) + I_{gl}$ under point + environment lighting, such that $I_{pt}^{(d)} + I_{env}$ is the direct component and $l_{tar} = [l_{pt}, l_{sh}]$. Moreover, the dataset lacks images under different lighting directions that are needed for relighting. Therefore, we follow \cite{sang2020single} to render target images under random point light positions from the frontal hemisphere in an online manner using $f_{BRDF}$ (see Equation \ref{eq:2}) and generate ground truth for supervision, allowing the model to learn from more samples under different lights. However, relighting is performed only for the direct component since $I_{gl}$ is only available for $l_{pt} = [0,0,0]$.

\textbf{Loss function.} We use L2 loss to supervise intrinsic components, image reconstruction, and relighting. Consider $\widehat{Y}$ as the estimate of the ground truth $Y$. The L2 loss $\mathcal{L}$ can be described as per Equation \ref{eq:6}.
\begin{equation}
    \centering
    \mathcal{L_{\star}} = \frac{1}{\sum_{i,j}M_{i,j}} ||(Y-\widehat{Y})\cdot M||_{2}^{2}
    \label{eq:6}
\end{equation}
Here, $Y \in \{A, R, N, D, I_{rec}, I_{rel}\}$ and $\mathcal{L}_{a}$, $\mathcal{L}_{r}$, $\mathcal{L}_{n}$, $\mathcal{L}_{d}$, $\mathcal{L}_{rec}$, and $\mathcal{L}_{rel}$ are the L2 losses for albedo, roughness,  normal, depth, reconstruction, and relighting. $M$ represents the object mask. The final loss function is given as follows.
\begin{equation}
    \centering
    \mathcal{L} = \lambda_{a}\mathcal{L}_{a} + \lambda_{n}\mathcal{L}_{n} + \lambda_{d}\mathcal{L}_{d} + \lambda_{r}\mathcal{L}_{r} + \lambda_{d}\mathcal{L}_{d} + \lambda_{rec}\mathcal{L}_{rec} + \lambda_{rel}\mathcal{L}_{rel}
\end{equation}
Here, $\lambda_{a} = \lambda_{r} = \lambda_{d} = \lambda_{rec} = \lambda_{rel} = 1.0$ and $\lambda_{n} = 2.0$. Additionally, we apply L2 loss over gradients of the roughness map. Mere L2 loss over roughness maps produces heavily flattened results and was observed to suppress specularities in the synthesized image. The reconstruction loss includes loss over direct and global illumination components as well. 

\textbf{Training Strategy.} We train our network end-to-end on \texttt{NVIDIA RTX 5000} with a batch size of $64$ using Adam optimizer \cite{kingma2014adam} with the initial learning rate of $1 \times 10^{-4}$ for image encoder and $2 \times 10^{-4}$ for decoders and global illumination network and decrease it by half after every five epochs for a total of $25$ epochs.

\begin{figure}[t]
	\centering
	\includegraphics[width=\textwidth]{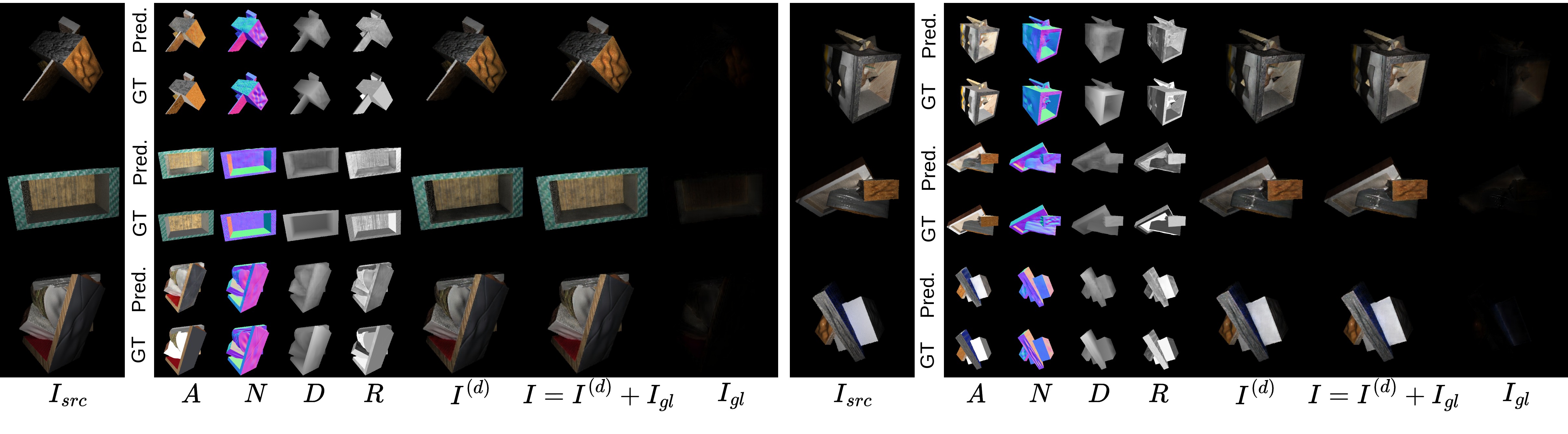}
	\caption{Qualitative results on the test set of \cite{li2018learning} emphasizing global-illumination effects. The superscript $(d)$ and subscript $(gl)$ represent the direct and global illumination components, respectively. Best viewed in PDF with zoom.}
	\label{fig:eval_res}

\end{figure}

\setlength{\tabcolsep}{4.0pt}
\begin{table}[h]
	\centering
	\caption{Quantitative results over different architectural design choices for images under point light sources from the test set of \cite{li2018learning}. Inp Img: whether the input image is under direct illumination (Img-d) or global illumination (Img-g). \#InvDec: the number of \textit{material} decoders. FS: Feature Sharing between \textit{relighting} and \textit{material} decoders, GA: Gated Attention, Rel: whether the relighting is through the neural network ($f_{rel}$) or directly through the BRDF model $(f_{BRDF})$, and GI: Global Illumination}
	\label{tab:1}
	\resizebox{\linewidth}{!}{%
		\begin{tabular}{c|cccccc|cccc|cc}
			\hline
			\multirow{2}{*}{ID} & \multicolumn{6}{c|}{Design Choices} & \multicolumn{4}{c|}{svBRDF Params (MSE $\times 10^{-2})$} & \multicolumn{2}{c}{Relighting (SSIM)} \\
			& Inp Img & \#InvDec & FS & GA & Rel & GI & A & R & N & D & Rel-$f_{rel}$ & Rel-$f_{BRDF}$  \\ \hline
			1 & Im-g & 1 & \ding{55} & \ding{55} & $f_{rel}$ & \ding{55} & 6.154 & 18.071  & 4.681 & 1.958 & 0.697 & 0.719 \\
			2 & Im-g & 1 & \ding{51} & \ding{55} & $f_{rel}$ & \ding{55} & 5.943  & 17.156 & 4.617  & 1.932 & 0.682 & 0.724 \\
                3 & Im-g & 1 & \ding{55} & \ding{51} & $f_{rel}$ & \ding{55} & 5.519 & 15.277 & 3.975 & 1.751 & 0.701 & 0.757 \\ 
                4 & Im-g & 1 & \ding{55} & \ding{55} & $f_{rel}$ & \ding{51} & 5.614 & 14.485 & 3.887 & 1.713 & 0.746 & 0.789 \\ 
			5 & Im-g & 1 & \ding{55} & \ding{51} & $f_{BRDF}$ & \ding{51} & 4.723 & 11.277 & 3.925 & 1.632 & - & 0.844 \\
			6 & Im-d & 1 & \ding{51} & \ding{51} & $f_{rel}$ & \ding{55} & 4.517 & 10.113 & 3.872 & 1.509 & 0.764  & 0.793 \\
			7 & Im-g & 1 & \ding{51} & \ding{51} & $f_{rel}$ & \ding{55} & 4.162 & 9.681 & 3.406 & 1.462  & 0.798 & 0.859 \\
			8 & Im-g & 4 & \ding{51} & \ding{51} & $f_{rel}$ & \ding{51} & \textbf{3.781} & 8.891 & 3.325 & 1.012  & 0.827 & 0.892 \\
			9 & Im-g & 1 & \ding{51} & \ding{51} & $f_{rel}$ & \ding{51} & 3.787  & \textbf{8.267} & \textbf{3.311} & \textbf{0.975} & 0.819 & \textbf{0.894} \\\hline
		\end{tabular}%
	}
\end{table}

\section{Experimental Evaluation}
\label{sec:result} 
 In this section, we compare the performance of MERLiN with benchmark methods \cite{li2018learning, sang2020single} on svBRDF estimation and relighting over synthetic and real data. Further, we demonstrate the closeness between the performance of photometric stereo benchmarks \cite{lichy2022fast, ikehata2023scalable} evaluated on relit images and their real counterparts.

\subsection{Ablation Studies}
Table \ref{tab:1} demonstrates the effect of several design choices quantitatively by evaluating the mean squared error obtained over the test set of \cite{li2018learning}. We observe that feature sharing across the two decoders of intrinsic parameter estimation and relighting performs better than without feature sharing (compare IDs $1$ and $2$, Table \ref{tab:1}). The same applies to components like gated attention (see IDs $1$ and $3$) and global illumination (see IDs $1$ and $4$). Interestingly, we observe that joint training helps improve svBRDF estimation, ensuring more physical correctness of inverse parameters (lower MSE for svBRDF params in ID $5$ and $9$). Note there is no feature sharing in experiment ID $3$ since there is no decoder for relighting. However, as shown in Figure \ref{fig:abl} (a), the images rendered using $f_{BRDF}$ are still better at capturing the global illumination effects such as specularities (higher SSIM under Rel-$f_{BRDF}$ across all the experiments). Therefore, we show the relighting results generated through $f_{BRDF}$ for all the experiments from hereon. The global illumination and gated attention perform better in tandem than excluding any of them (compare IDs $7$ and $9$). We also observed that under similar settings, using one decoder for svBRDF parameter estimation offers near-close performance compared to four explicit decoders (one for each parameter) except for the albedo (IDs $8$ and $9$). This marginal under-performance could be acceptable for reduced network size and higher run-time speed. Moreover, taking images with global illumination as input produces far better results and generalizes well to real images than images under direct illumination (see IDs $6$ and $9$ and Figure \ref{fig:abl}: Row 3). Firstly, images under only direct illumination seldom exist in the real world. Moreover, they offer limited information when dealing with arbitrary shapes and materials.

\setlength{\tabcolsep}{15pt}
\begin{table}[t]
	\centering
	\caption{Quantitative comparison of svBRDF estimation (MSE $\times 10^{-2}$) and relighting (SSIM) of MERLiN with Li \emph{et al.} \cite{li2018learning} and Sang \emph{et al.} \cite{sang2020single} over images under point light global illumination from the test set of \cite{li2018learning}. }
	\label{tab:2}
	\resizebox{\textwidth}{!}{%
		\begin{tabular}{c|cccc|c}\hline
			Method & A & R & N & D & Relighting \\\hline
			Li \emph{et al.} \cite{li2018learning} & 4.868 & 19.431 & 3.822 & 1.505 & 0.884  \\
			Sang \emph{et al.} \cite{sang2020single} & 3.856 & 12.781 & 3.459 & 1.471 & 0.872    \\
			MERLiN (Ours) & \textbf{3.787} & \textbf{8.267} & \textbf{3.311} & \textbf{0.975} & \textbf{0.894}   \\\hline
		\end{tabular}%
	}

\end{table}
\subsection{Quantitative Results on svBRDF Estimation, Reconstruction, and Relighting}
We evaluate and compare MERLiN with the two closest benchmark methods \cite{li2018learning} and \cite{sang2020single} over the test set of \cite{li2018learning} in Table \ref{tab:2}.  We obtained significantly improved svBRDF parameter estimation results over both methods, along with image reconstruction and relighting. It is important to note that while \cite{li2018learning} has been trained on global illumination images, \cite{sang2020single} considers direct illumination without explicitly considering global illumination effects. However, both frameworks are evaluated on images with global illumination. As a result, we see reduced relighting performance of \cite{sang2020single} compared to \cite{li2018learning}. Interestingly, our in-network end-to-end training of the global illumination layer proves better than the ad-hoc training of the global illumination network in \cite{li2018learning} over relighting.
\begin{figure}[t]
	\centering
	\includegraphics[width=\textwidth]{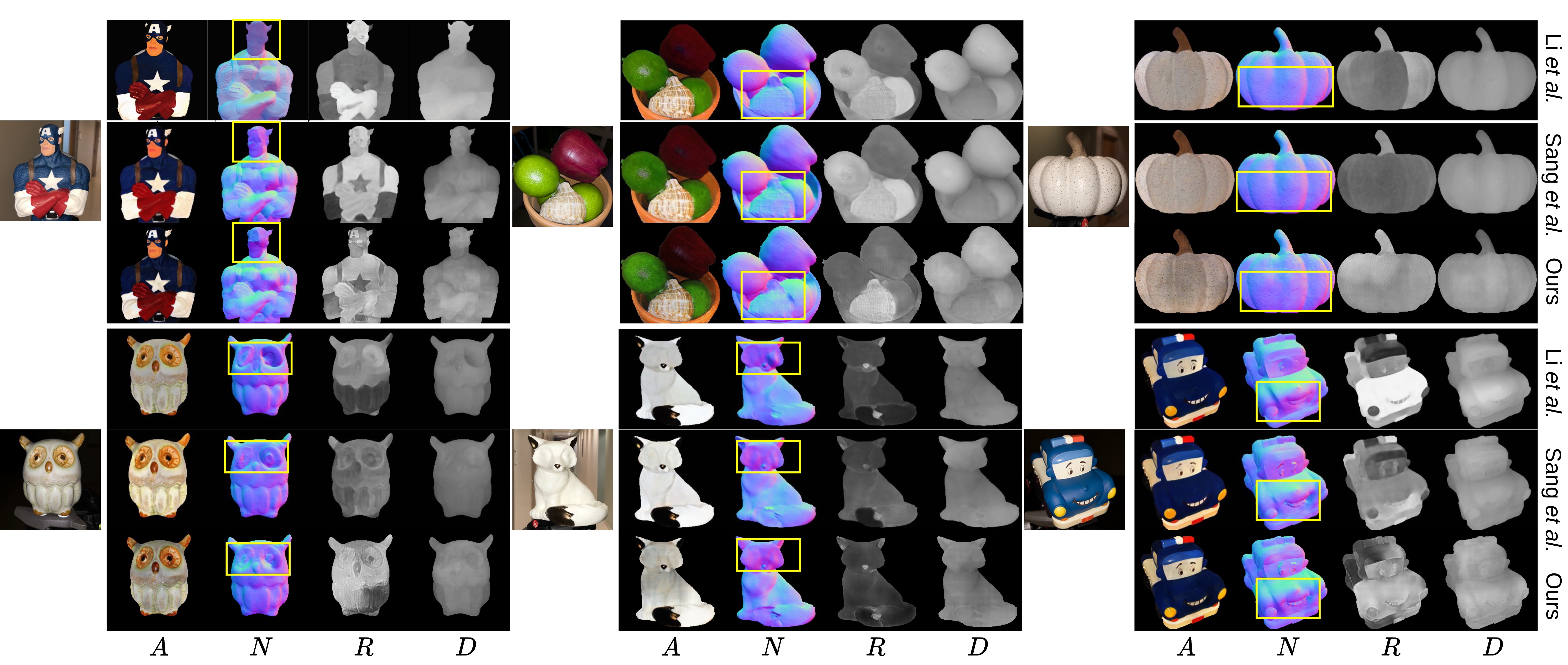}
	\caption{Qualitative comparison of svBRDF parameters: albedo (A), normal (N), roughness (R), and depth (D) among MERLiN and methods \cite{sang2020single} and \cite{li2018learning}. Differences can be observed in the marked regions across different svBRDF parameters.}
	\label{fig:brdf}
  
\end{figure} 

\begin{figure}[h]
	\centering
	\includegraphics[width=\textwidth]{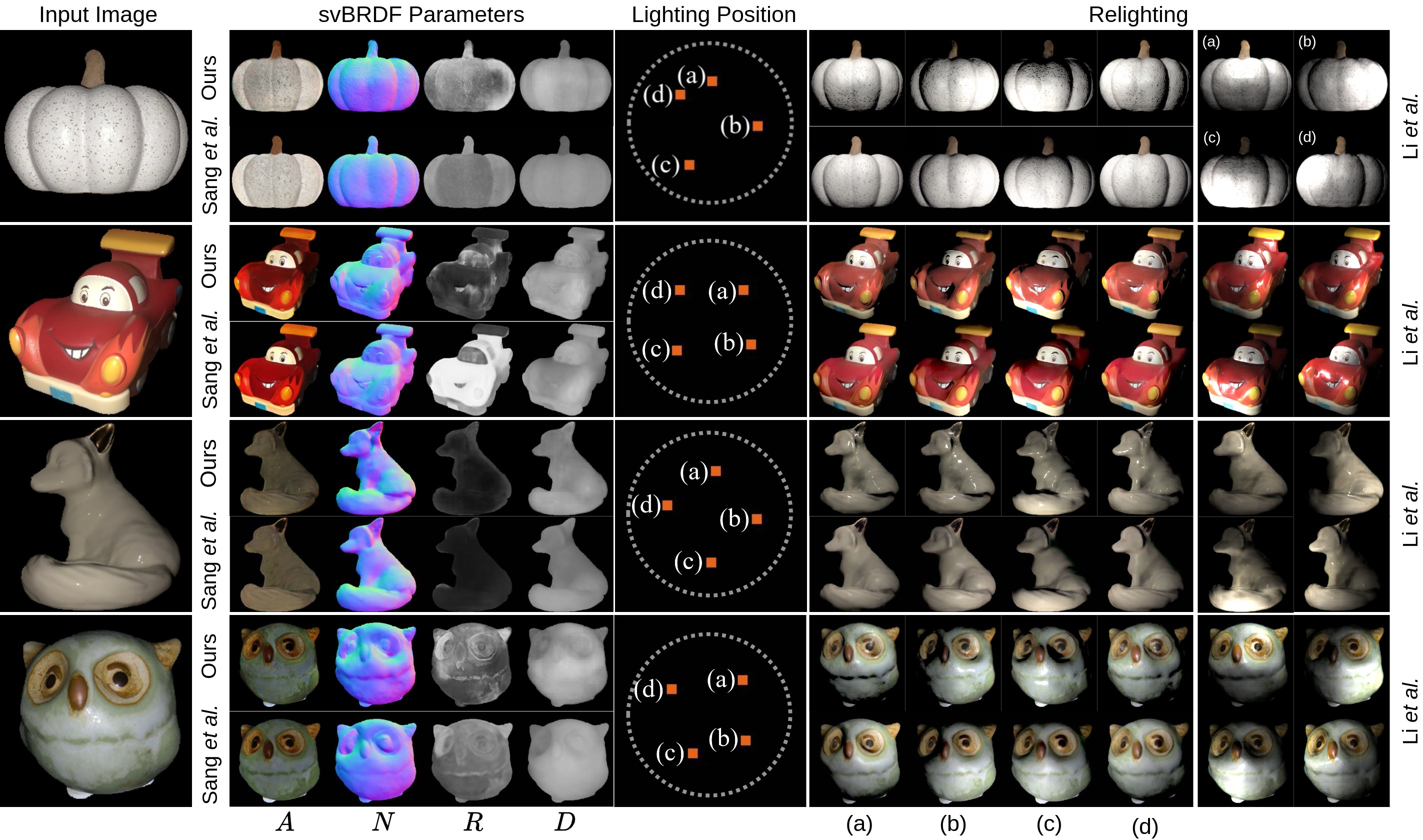}
	\caption{Qualitative evaluation of the relit images generated through MERLiN, \cite{sang2020single}, and \cite{li2018learning} under point lighting over the test dataset of real images.}
	\label{fig:brdf_rel_res}

\end{figure}

\subsection{Qualitative Results} 
\begin{figure}[t]
	\centering
	\includegraphics[width=\linewidth]{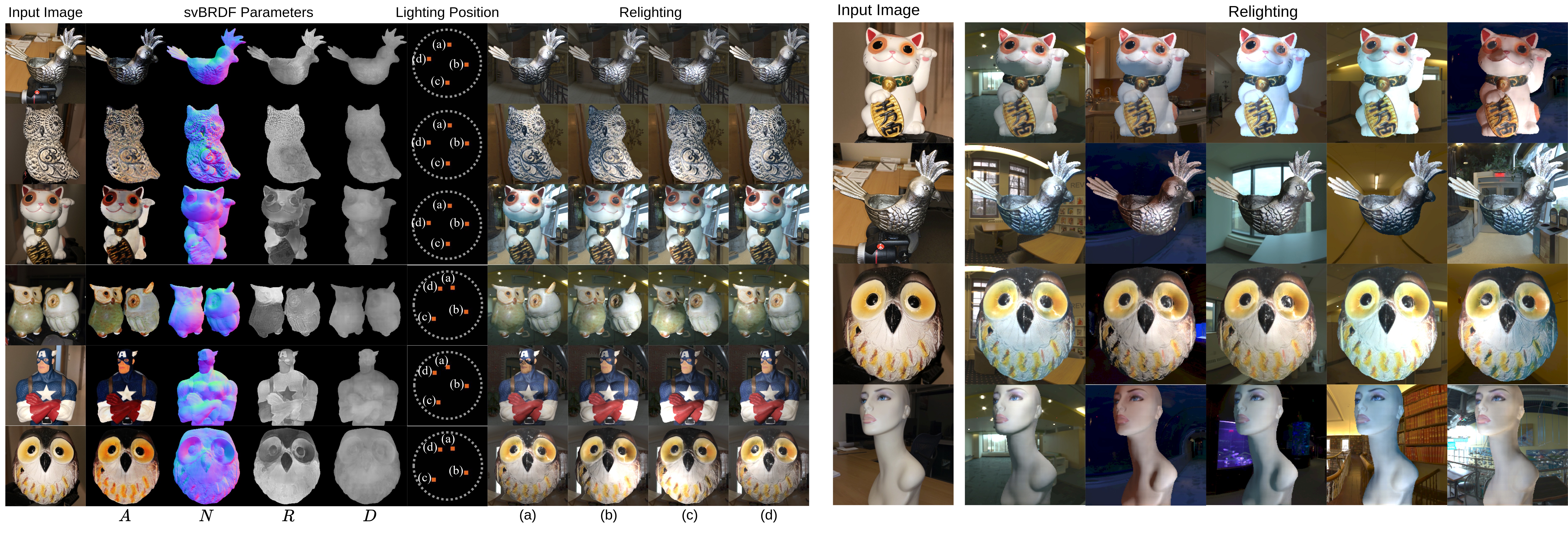}
	\caption{Relighting through MERLiN under arbitrary environment lightings.}
	\label{fig:env_res}

\end{figure}

We perform extensive qualitative evaluation over synthetic and real data, especially due to the lack of ground truth for quantitative comparison. Figure \ref{fig:eval_res} shows the results over the test set of \cite{li2018learning}. While the estimated svBRDF parameters are close to ground truth, the network also reasonably captures the global illumination effects. The proposed residual design does not explicitly model inter-reflections from surface points that are not visible to the camera. Instead, it only approximates the true global illumination by operating in image space and learning from inter-reflections in the training data. Figure \ref{fig:brdf} shows the results over real images. We observe that MERLiN produces better inverse parameters than \cite{li2018learning} and \cite{sang2020single} just from a single-stage architecture. Since physically plausible relighting is the key to our work, we compare the relit images across four objects in Figure \ref{fig:brdf_rel_res}. While \cite{sang2020single} fails to model sharp surface specularities, mainly due to its inability to handle global illumination explicitly. Moreover, this observation aligns with the finding that neural network-based relighting (even with joint training) suffers in handling the global illumination effects but helps in better inverse parameter estimation. The relighting using $f_{rel}$ in MERLiN (see Figure \ref{fig:abl} (a)) produces similar results as that of \cite{sang2020single}. Moreover, \cite{li2018learning} models the specularities a little better, but it spreads the specularities over a larger area. Overall, MERLiN obtains better relighting results exhibiting high photorealism consistent with the underlying material parameters. Figure \ref{fig:env_res} shows estimated inverse parameters and relighting under environment lighting by varying the point light and the environment map. We observe that the network generates highly realistic images under arbitrary environments. 

\begin{SCfigure}[1.0][h]
	\centering
	\includegraphics[width=0.675\textwidth]{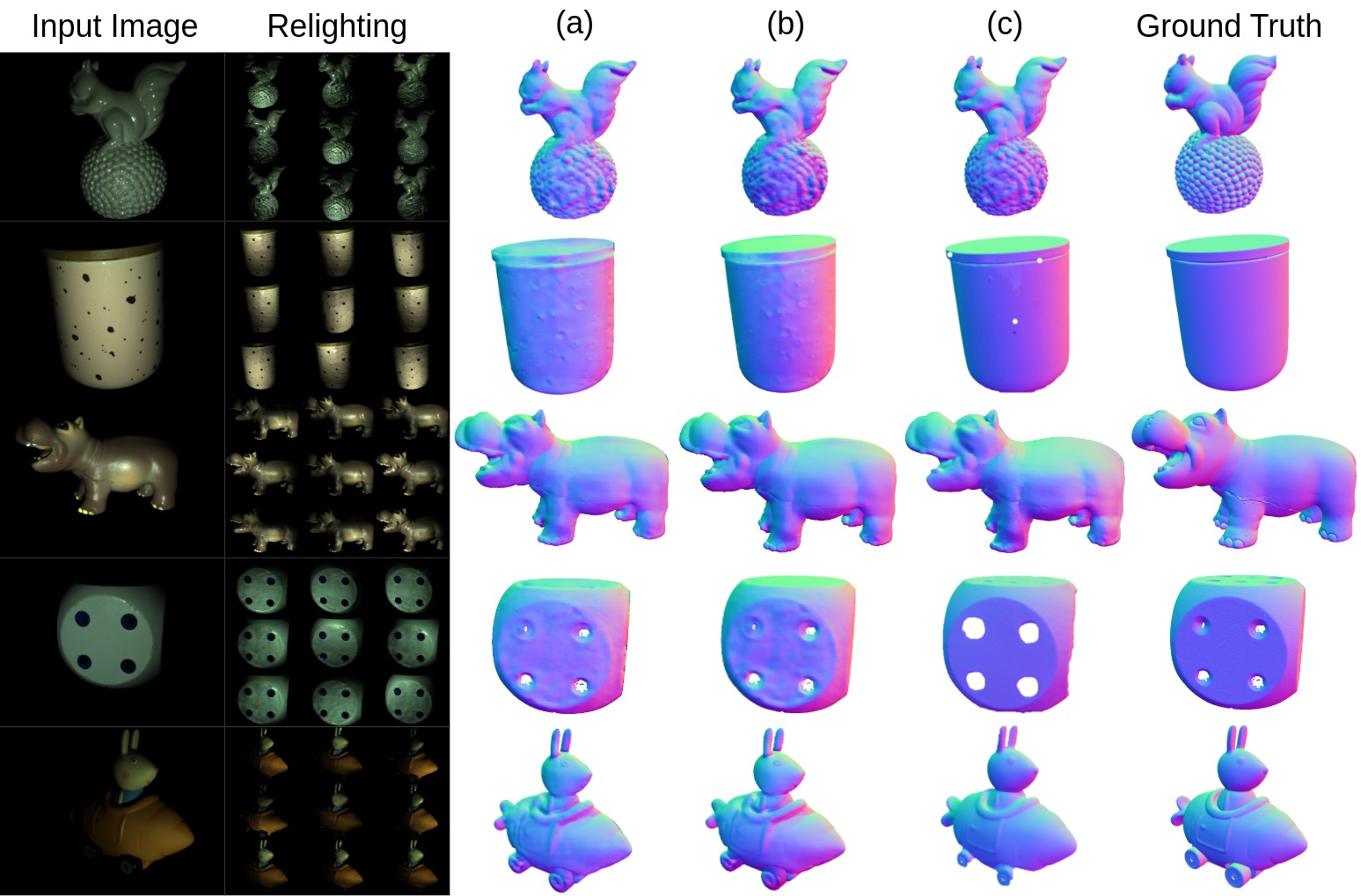}
	\caption{Qualitative results on photometric stereo over LUCES dataset \cite{mecca2021luces} We compare (a) single image-based normals obtained by the material decoder of MERLiN and the normals estimated by Fast-NFPS \cite{lichy2022fast} through (b) relit images by MERLiN and (c) real 32 images. Note that the relit images are generated by MERLiN using a single image.}
	\label{fig:pps}
\end{SCfigure}

\begin{figure}[h]
	\centering
	\includegraphics[width=\linewidth]{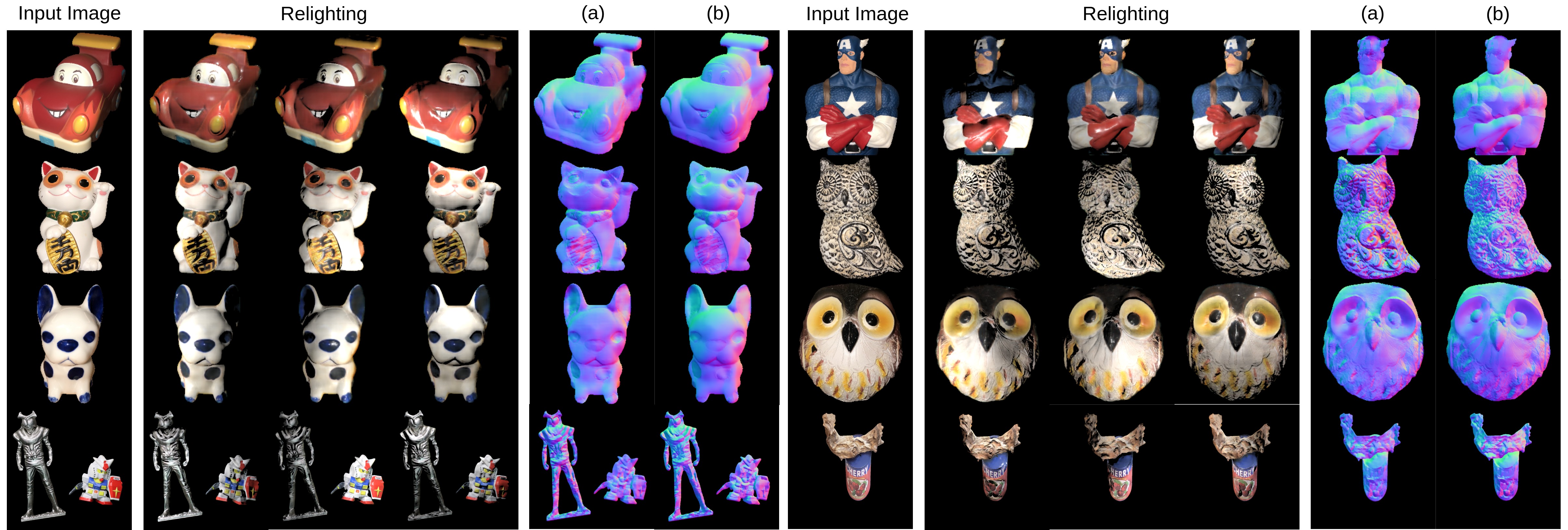}
	\caption{Qualitative results on photometric stereo over real test dataset \cite{sang2020single}.  We compare (a) single image-based normals obtained by the material decoder of MERLiN and the normals estimated by SDM-UniPS \cite{ikehata2023scalable} through (b) relit images by MERLiN.}
	\label{fig:ps}
\end{figure}

\section{Photometric Stereo through Relit Images} Once we have synthesized the images under arbitrary lighting from a single input image, we feed them through the two classic benchmark photometric stereo methods - Fast-NFPS \cite{lichy2022fast}  and SDM-UniPS \cite{ikehata2023scalable} to evaluate the estimated normal maps. We examine their performance over the LUCES dataset \cite{mecca2021luces} - a photometric dataset under near-field lighting. Since MERLiN is trained on images with the perspective camera and near-field lights, evaluation over Fast-NFPS using synthetically relit images from the LUCES dataset is best suited for us. 

The rationale behind evaluating photometric stereo over relit images is to answer three questions. (a) Are the relit images physically correct? (b) Are the normal estimates using multiple relit images better than those from a single image? (c) How close are the results when compared to their real counterparts? 

In an attempt to answer these questions, we perform a quantitative comparison of the photometric stereo performance using the Fast-NFPS method \cite{lichy2022fast} over relit images generated by MERLiN, Sang \emph{et al.} \cite{sang2020single}, and Li \emph{et al.} \cite{li2018learning}. Table \ref{tab:3} reports the mean angular error (MAE) in degrees over the estimated normals. Specifically, we select an image under near frontal lighting (closest to the camera-light collocated setup) from the LUCES dataset and generate $50$ images under lighting from the frontal hemisphere using all the three methods MERLiN, \cite{sang2020single}, and \cite{li2018learning}. For each method, we randomly select $32$ images from the set of $50$ images over $5$ different runs and pass them through Fast-NFPS under an uncalibrated setting to observe the final MAE. 
Our quantitative analysis shows that while Fast-NFPS achieves lower MAE over real images compared to relit images generated by MERLiN, the performance surpasses that of normal estimates derived from a single image through MERLiN and the other relighting methods \cite{sang2020single, li2018learning}. Furthermore, we observe that Sang \emph{et al.} \cite{sang2020single} performs well on relatively flat object surfaces but produces higher MAE over complex objects like Buddha, House, and Squirrel, owing to its inability to handle the cast shadows and indirect illumination arising out of underlying surface variations. We also show the qualitative results in Figure \ref{fig:pps} comparing the single image normals with normals through relit and real images. We observe that the multiple relit images offer better, if not the same, normal estimates compared to the single image. We also evaluate the SDM-UniPS performance qualitatively on relighted images using the samples from real test data of \cite{li2018learning, sang2020single}. SDM-UniPS is an interesting choice as it learns global lighting context and is agnostic to any physical lighting model. Figure \ref{fig:ps} shows the qualitative comparison of normals estimated by a single image using MERLiN and SDM-UniPS through relit images generated using MERLiN. The photometric stereo through relit images indeed improves the normal estimation, especially evident in Figure \ref{fig:ps}: rows $2$ and $3$. However, we find that the method gets distracted by image semantics (picture of cherry on the can, Figure \ref{fig:ps}, last row) to provide erroneous surface normals.

\setlength{\tabcolsep}{1pt}
\begin{table}[t]
	\centering
	\caption{Mean angular error (MAE) over the normals estimated through a single image, and sets of real and relit images under $32$ different point lighting from the LUCES dataset \cite{mecca2021luces} over Fast-NFPS \cite{lichy2022fast} method.}
	\label{tab:3}
	\resizebox{\textwidth}{!}{%
		\begin{tabular}{c|c|cccccccccccccc|c}\hline
			Input  & Rel. Method & Bell & Ball & Buddha & Bunny & Die & Hippo & House & Cup & Owl & Jar & Queen & Squirrel & Bowl & Tool & Average \\\hline
			Single Image & -
			& 12.03 & 10.75 & 21.26 & 12.02 & 9.51 & 11.23 & 40.16 & 19.68 & 17.62 & 9.37 & 20.93 & 19.94 & 12.79 & 21.59 & 17.06 \\ \hline
			\multirow{3}{*}{$32$ Relit Images} & Sang \emph{et al.} \cite{sang2020single} 
		  & 10.09 & 9.52 & 19.17 & 12.69 & 9.21 & 10.08 & 39.42 & 19.59 & 17.29 & 9.79 & 22.19 & 19.67 & 11.96 & 19.29 & 16.43 \\
             & Li \emph{et al.} \cite{li2018learning}  
        	& 10.33 & 9.89 & 18.96 & 12.03 & 10.04 & 10.11 & 36.88 & 19.34 & 16.17 & 10.51 & 21.31 & 19.32 & 12.23 & 19.77 & 16.21 \\
			& MERLiN (Ours)
			& 9.51 & 9.12 & 18.27 & \textbf{11.71} & 9.12 & \textbf{10.02} & 36.91 & 19.27 & 16.97 & 9.82 & 20.18 & 19.05 & 11.98 & 19.31 & 15.80 \\\hline
            $32$ Real Images  & -
			& \textbf{7.17} & \textbf{6.59} & \textbf{14.50} & 11.89 & \textbf{8.63} & 10.64 & \textbf{31.00} & \textbf{18.98} & \textbf{15.92} &\textbf{ 9.14} & \textbf{18.39} & \textbf{18.26} & \textbf{10.17} & \textbf{18.61} & \textbf{14.11} \\\hline
		\end{tabular}%
	}
\end{table}

\section{Conclusion}
\label{sec:conclusion}
In this work, we took a step towards addressing the data acquisition challenge in photometric stereo through joint material estimation and relighting from a single image. Our single-stage MERLiN network outperformed the baselines with cascaded network architectures over material estimation and relighting, offering faster run-time speed and a low memory footprint. Moreover, explicit global illumination rendering proved effective across all the experiments. Further, we evaluated photometric stereo methods over relit images synthesized from a single input image, shedding light on key questions regarding the physical correctness of relit images, the efficacy of normal estimation using multiple relit images compared to a single image, and the fidelity of results when compared to the real counterparts.   It can even be applied to dynamic surface recovery, where a single instance of a dynamic surface can be analyzed under different lighting, allowing photometric stereo for dynamic surfaces.

\section*{Acknowledgements}
We would like to acknowledge the generous support of the Prime Minister Research Fellowship (PMRF) grant, the Japan Society for the Promotion of Science (JSPS) KAKENHI Grant (Number JP24K02966), and the Jibaben Patel Chair in Artificial Intelligence for the completion of this work.

%
%
\bibliographystyle{splncs04}
\bibliography{main}
\end{document}


\title{MERLiN: Single-Shot Material Estimation and Relighting for Photometric Stereo\\(Supplementary)} 

\titlerunning{MERLiN}

\author{Ashish Tiwari\inst{1}\orcidlink{0000-0002-4462-6086} \and
Satoshi Ikehata\inst{2}\orcidlink{0000-0002-6061-7956} \and
Shanmuganathan Raman\inst{1}\orcidlink{0000-0003-2718-7891}}

\authorrunning{A.~Tiwari et al.}

\institute{Indian Institute of Technology Gandhinagar, Gujarat, India \\ 
\email{\{ashish.tiwari, shanmuga\}@iitgn.ac.in}\\ \and
National Institute of Informatics, Tokyo, Japan\\
\email{sikehata@nii.ac.jp}}

\maketitle

\section{Relighting Results}
We have compiled the results on relighting in a video showing the comparison between Li \emph{et al.} \cite{li2018learning}, Sang \emph{et al.} \cite{sang2020single}, and MERLiN (Ours).  We show the results under point lighting, environment lighting, and point $+$ environment lighting. We observe the change of specular highlights when rotating the point light source and the environment maps, which shows that spatially varying roughness has been successfully captured by the network. Moreover, we observe that Li \emph{et al.} \cite{li2018learning} models the specularities slightly better than Sang \emph{et al.} \cite{sang2020single}, but it spreads the specularities over a larger area and also produces saturating effects (as shown in the regions highlighted in the green box). The relit images generated by MERLiN appear more realistic and physically plausible. We do not have ground truth images under respective lighting for real data. However, the realistic changes in shading and specular highlights while rotating light sources could be observed to appreciate the relighting performance.

\noindent The video can be found at: \url{https://sites.google.com/iitgn.ac.in/merlin}

\section{BRDF Model}

We use the microfacet BRDF model described in  \cite{karis2013real}. Given the diffuse albedo (A), specular roughness (R), normals (N), and depth (D) along with $l$, $v$, and $h$ being the light direction, viewing angle, and the half-angle between them, the BRDF $(f_{BRDF})$ can be characterized as per Equation \ref{eq:3}.
\begin{equation}
    \centering
    f_{BRDF}(A,N,R,D,l,v) = \frac{A}{\pi} +  \frac{\widetilde{M}_{f}(h, R)\widetilde{F}(v,h)\widetilde{G}(l,v,h,R)}{4(N\cdot l)(N \cdot v)}
    \label{eq:3}
\end{equation}
Here, $\widetilde{M}_{f}(h,R)$, $\widetilde{F}(v,h)$, and $\widetilde{G}(l,v,h,R)$ are the microfacet distribution, Fresnel, and geometry term. The distribution term describes the distribution of surface normals representing the probability density function of the microfacet normals oriented along the direction. The Fresnel term accounts for how much light is reflected versus refracted at the interface between two media, and the geometry term accounts for occlusion between microfacets.

Each term in Equation \ref{eq:3} is defined as follows.

\begin{equation*}
    \centering
    \widetilde{M}_{f}(h, R) = \frac{\alpha^{2}}{\pi [(N \cdot h)^{s}(\alpha^{2}-1) + 1]^{2}}
\end{equation*}

\begin{equation*}
    \centering
    \alpha = R^{2}
\end{equation*}

\begin{equation*}
    \centering
    \widetilde{F}(v, h) = (1 -F_{0})2^{-[5.55473(v\cdot h) + 6.8316] (v \cdot h)}
\end{equation*}

\begin{equation*}
    \centering
    \widetilde{G}(l,v,h,R) = \widehat{G}_{1}(v, N) \widehat{G}_{1}(l,N)
\end{equation*}
\begin{equation*}
    \centering
    \widetilde{G}_{1}(v,N) = \frac{N \cdot v}{(N \cdot v)(1 - k) + k}
\end{equation*}
\begin{equation*}
    \centering
    \widetilde{G}_{1}(l,N) = \frac{N \cdot l}{(N \cdot l)(1 - k) + k}
\end{equation*}
\begin{equation*}
    \centering
    k = \frac{(R+1)^{2}}{8} 
\end{equation*}
\begin{equation*}
    \centering
    F_{0} = 0.05 
\end{equation*}

\bibliographystyle{splncs04}
\bibliography{main}